\documentclass{article}

\usepackage{microtype}
\usepackage{graphicx}
\usepackage{subfigure}
\usepackage{booktabs} 

\usepackage{amsmath}
\usepackage{etoolbox}

\usepackage{amssymb}
\usepackage{algorithmic}
\usepackage{forloop}
\usepackage{fancyhdr}
\usepackage{stmaryrd}
\usepackage{textcomp}
\usepackage{bm}

\usepackage{amssymb}
\usepackage{booktabs}
\usepackage{multirow}
\usepackage{kotex}
\usepackage{tablefootnote}
\usepackage{footnote}
\usepackage[dvipsnames,table,xcdraw]{xcolor}
\usepackage{lineno}
\usepackage{hyperref}
\usepackage{graphicx}
\usepackage{ulem} 
\usepackage{textcomp}
\usepackage{latexsym}
\usepackage{graphicx}
\usepackage{tabularx}
\usepackage{graphics}
\usepackage{adjustbox}
\usepackage{algorithmic}
\graphicspath{{./figures/}}

\usepackage{relsize}
\usepackage{mathtools}
\usepackage{hyperref}

\usepackage[accepted]{icml2021}

\icmltitlerunning{LIVENESS SCORE-BASED REGRESSION NEURAL NETWORKS FOR FACE ANTI-SPOOFING}

\begin{document}

\twocolumn[
\icmltitle{LIVENESS SCORE-BASED REGRESSION NEURAL NETWORKS FOR FACE ANTI-SPOOFING}



\icmlsetsymbol{equal}{*}

\begin{icmlauthorlist}
\icmlauthor{Youngjun Kwak}{kakao,kaist}
\icmlauthor{Minyoung Jung}{airc}
\icmlauthor{Hunjae Yoo}{kakao}
\icmlauthor{JinHo Shin}{kakao}
\icmlauthor{Changick Kim}{kaist}
\end{icmlauthorlist}

\icmlaffiliation{kakao}{KakaoBank Corp., South Korea}
\icmlaffiliation{airc}{AIRC, Korea Electronics Technology Institute, South Korea}
\icmlaffiliation{kaist}{Department of Electrical Engineering, KAIST, South Korea}

\icmlcorrespondingauthor{Changick Kim}{changick@kaist.ac.kr}

\icmlkeywords{Face anti-spoofing, Regression neural network, Label encoding, Bio-metrics}

\vskip 0.3in ]



\printAffiliationsAndNotice{}  

\begin{abstract}
Previous anti-spoofing methods have used either pseudo maps or user-defined labels, and the performance of each approach depends on the accuracy of the third party networks generating pseudo maps and the way in which the users define the labels. In this study, we propose a liveness score-based regression network for overcoming the dependency on third party networks and users. 
First, we introduce a new labeling technique, called pseudo-discretized label encoding for generating discretized labels indicating the amount of information related to real images.
Secondly, we suggest the expected liveness score based on a regression network for training the difference between the proposed supervision and the expected liveness score.
Finally, extensive experiments were conducted on four face anti-spoofing benchmarks to verify our proposed method on both intra-and cross-dataset tests. The experimental results denote our approach outperforms previous methods.
\end{abstract}

\vspace{-1.1em}
\section{Introduction}
\label{sec:intro}

\begin{figure}[]
  \centering
  {\includegraphics[width=\columnwidth]{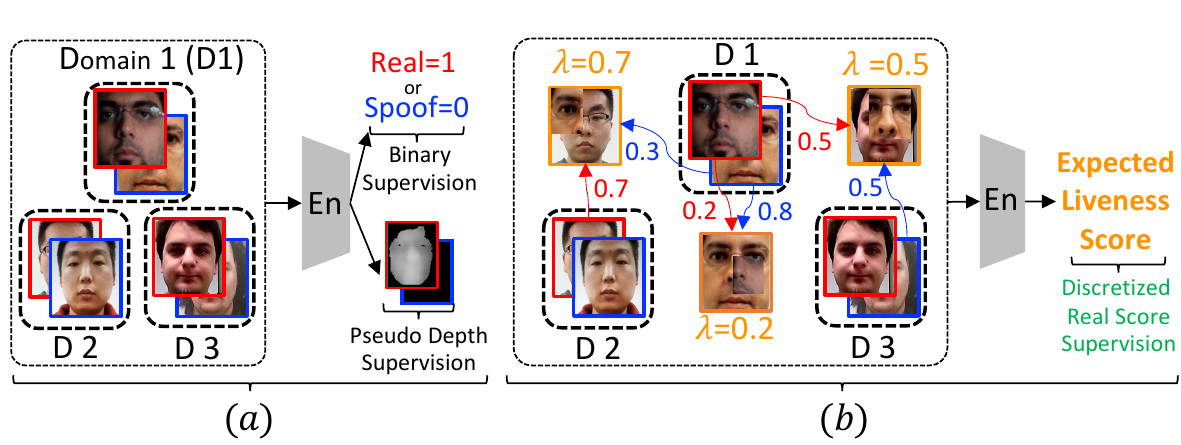}}\vspace{-1.3em}
\caption{Comparison between previous methods and our method for face anti-spoofing. (a) Previous methods utilize either binary supervision to detect spoof cues, or pseudo depth supervision, or both. (b) Our method discretizes binary labels and exchanges real and spoof images for our expected liveness score. The discretized label $\lambda$ indicates the ratio of a real image over an image.}
\vspace{-1.0em}
\label{fig:fig1}
\end{figure}

\begin{figure*}[ht!]
\includegraphics[width=\textwidth]{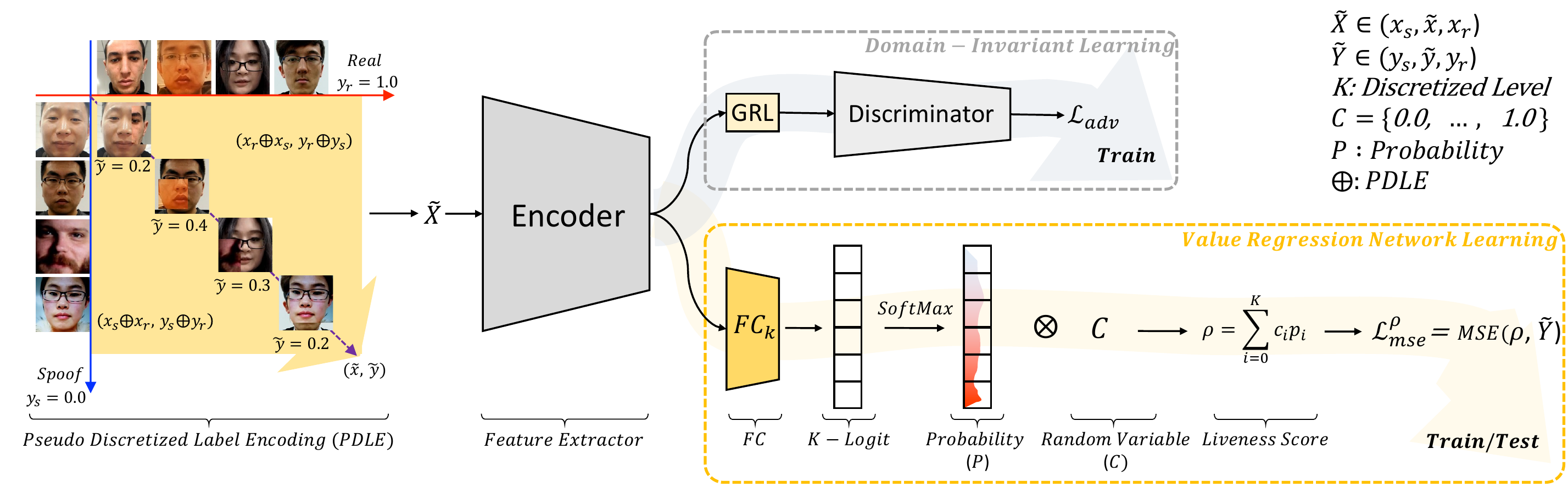}\\[-0.1pt]
\vspace{-2.5em}
\caption{Overview of our approach for a value regression neural network. Our framework consists of a label encoding (PDLE) for the data and label expansion, a encoder network for the feature extractor, an expected liveness score estimator for the regression network learning, and a discriminator for the domain-invariant feature learning.}
\vspace{-0.3em}
\label{fig:fig2}
\end{figure*}

Face anti-spoofing (FAS) systems have been successfully established in face authentication, and widely used in online banking, electronic payments, and securities as a crucial technique. Despite its substantial success, FAS still shows vulnerability to various presentation attacks (PAs) such as printed materials, replay-videos, and 3D-masks. To alleviate such vulnerability, previous deep learning-based FAS methods ~\cite{Liu_2018_CVPR, yu2020searching} learn discriminative features for distinguishing real faces against PAs, and such methods mostly treat the FAS problem as a binary classification of whether a given face is real or a spoof, as shown in Fig.~\ref{fig:fig1}(a).
However, such binary classification-based approaches suffer from non-trivial attacks because they are prone to an over-fitting to the training data, resulting in poor generalization~\cite{Liu_2018_CVPR}. To mitigate the over-fitting problem, regression-based methods~\cite{feng2018prn,MEGC,yu2021dual,yu2020searching} have been proposed, which find sparse evidence for known spoof types and generalize to unseen attacks. For regression-based neural networks, two approaches are considered:
First, pseudo-define based supervision~\cite{DBEL,MEGC,yu2021dual,yu2020searching,Liu_2018_CVPR,fgmask} is designed for context-agnostic discrimination describing the local cues from the pixel level, such as the depth and reflection.
For example, a pseudo-map based CNN~\cite{feng2018prn} utilizes pseudo-depth supervision using the mean square error (MSE) to reconstruct a sparse depth-map and a flattened-map for a real and spoof image, respectively, as illustrated in Fig.~\ref{fig:fig1}(a). And PAL-RW~\cite{fgmask} is the first approach that utilizes partial pixel-wise labels with face masks to train FAS models. Secondly, user-define based supervision~\cite{ordinalreg, Wang_2022_CVPR} is designed for constrained learning using the relative distances among real and PAs to improve the generalization ability. For instance, ordinal regression~\cite{ordinalreg} introduces user-defined ordinal labels. Based on the user-defined labels, the model is trained to finely constrain the relative distances among the features of different spoof categories within the latent space. Another example is PatchNet~\cite{Wang_2022_CVPR}, which subdivides binary labels (a real or a spoof) into fine-grained labels (reals or spoofs). Despite previous efforts, we found that the pseudo-define based supervisions depend on the accuracy of additional works (e.g., depth-~\cite{feng2018prn} and texture-based~\cite{zhang2018single}),
and that user-define based supervision relies on that user-specified guides, and the correctness is not guaranteed. In this paper, as described in Fig.~\ref{fig:fig1}(b), we introduce a discretized label encoding for increasing data distribution and generating data relationships, which has no dependencies on the prior works. For our proposed label encoding, we present a novel pre-processing method, called the pseudo-discretized label encoding (PDLE) scheme, in which an image is randomly selected in a mini-batch, then the opposite labeled image is also arbitrarily chosen from the whole batch, and then parts of the images are exchanged to generate a new image and its discretized dense label.

Our contributions are as follows:
\vspace{-1.1em}
\begin{itemize}
\item We re-formulate face anti-spoofing as a value regression problem that directly optimizes a deep neural network with mean square error for improving performance, instead of using binary cross-entropy.
\item We propose a simple yet effective pseudo-discretized label encoding (PDLE), which enforce the regression network to represent the ratio of information of the real image to that of the given input image for a prediction of the liveness score.
\item We conduct extensive experiments, and obtain the state-of-the-art and outstanding performance on the intra- and cross-dataset respectively.
\end{itemize}

\vspace{-1.5em}
\section{Proposed Method}
\label{sec:method}
\vspace{-0.5em}
\subsection{Overview}
\vspace{-0.5em}
For an expansion of the training image and label distribution without an information corruption, we introduce a discretized label encoding schema to preserve the spoof and real cues in the images, and indicate the amount of a real image information over that of the input image. To leverage the PDLE, we propose learning a value regression neural network using the MES between the expected liveness scores and the pseudo-labels. In addition, we apply a domain-invariant learning scheme (GRL)~\cite{ganin2015unsupervised} as an adversarial training to our regression neural network using the domain labels. The framework of our method is illustrated in Fig.~\ref{fig:fig2}.

\vspace{-0.5em}
\subsection{Pseudo-Discretized Label Encoding}
\vspace{-0.5em}
We assume that $X=\{x_{s},x_{r}\}\in \mathbb{R}^{H\times W\times 3}$ and $Y=\{y_{s}=0.0,y_{r}=1.0\}$ denote the spoof and real color image space and the class label space in each. To sample the discretized labels between $y_{s}$ and $y_{r}$, we use the following formula:

\vspace{-1.2em}
\begin{equation}
\begin{aligned}
    u \sim \mathcal{U}\{1, K\}; \ \ \ \ \lambda = \frac{u}{K},
\end{aligned}
\label{eq:eq1}
\end{equation}
where $u$ is sampled from the discrete uniform distribution (1, $K$), and $K$ is a pre-defined discretized level, a cardinality of an encoding label $\tilde{Y}$ set, and a number of outputs for the last $FC$ in Fig.~\ref{fig:fig2}. $\lambda$ implies a pseudo-discretized label presenting the amount of a partial real image over a whole image. Inspired by CutMix~\cite{yun2019CutMix}, we first exchange a real image and a spoof image through a random rectangular box as follows:

\vspace{-1.2em}
\begin{equation}
\begin{aligned}
    \tilde{x} &= M\odot x_{a} + (1-M)\odot x_{b}, \text{where } y_{a} \neq y_{b}\\ 
    \tilde{y} &= 
\begin{cases}
    1-\lambda,& \text{if } x_{a} = x_{r}\\
    \lambda,              & \text{otherwise},
\end{cases}
\end{aligned}
\label{eq:eq3}
\end{equation}
where $M \in \{0,1\}^{H\times W}$ is a random rectangular mask based on $\lambda$, with $0$ and $1$ indicating inside and outside the mask. $\odot$ is an element-wise multiplication operator, and $x_{a}$ is an anchor to choose a sample from a mini-batch, whereas $x_{b}$ is the opposite sample selected from the entire training set. $\tilde{x}$ indicates the exchanged image, and $\tilde{y}$ is the pseudo-discretized label determined based on whether $x_{a}$ is a real image or not. We exchange between images with different labels ($y_{a} \neq y_{b}$) to expand data and label distribution. As shown in Fig.~\ref{fig:fig2}, we use $\tilde{X} \in (x_{s}, \tilde{x}, x_{r})$ and $\tilde{Y} \in (y_{s}, \tilde{y}, y_{r})$ as the training data and the supervision for the regression network to learn the liveness score.

\vspace{-1.0em}
\subsection{Expected Liveness Score}
\vspace{-0.5em}
Let $\mathbb{P}:\mathbb{R}^{H\times W \times 3} \rightarrow \mathbb{R}^{K}$ denote the probability of the liveness evidence estimated using
$SoftMax$, $FC_{k}$, and $Encoder(\tilde{X})$, as illustrated in Fig.~\ref{fig:fig2}. We employ $K$ in Eq.~\ref{eq:eq1} to formulate a random variable $C$ with a finite list $\{c_{0}, ..., c_{K}\}$ whose the $i^{th}$ element $c_{i}$ is denoted as follows:


\vspace{-1.5em}
\begin{equation}
\begin{aligned}
    c_i &= 
    \begin{cases}
    0.0,& \text{if } i = 0\\
    interval\times i ,& \text{if } i > 0 \text{ and } i < K\\
    1.0,& \text{if } i = K\\
\end{cases}
\end{aligned}
\label{eq:eq4}
\vspace{-0.5em}
\end{equation}
where $interval=\lceil \frac{y_{r} - y_{s}}{K}\rceil$. The random variable $c_{i}$ and its probability $p_{i}$ are exploited to calculate the expected liveness score as follows:

\vspace{-1.8em}
\begin{equation}
\mathbb{E}[C] = \rho = \sum_{i=0}^{K} c_{i}*p_{i},
\label{eq:eq5}
\vspace{-0.5em}
\end{equation}
where $p_{i}$ is the $i^{th}$ element of $P$ which is the predicted probability vector of real cues from the input $\tilde{X}$. We write $\mathbb{E}[C]$ with $\rho$, which is calculated using the sum over the element-wise multiplication between the random variables and their corresponding probabilities.

\vspace{-0.5em}
\subsection{Objective Function}
\vspace{-0.5em}
Our objective function is defined as follows:

\vspace{-1.0em}
\begin{equation}
L^{\rho}_{mse} = -\frac{1}{N}\sum_{j=1}^{N}(\rho_{j} - \tilde{Y}_{j})^{2},
\label{eq:eq6}
\end{equation}
where $N$ is a mini-batch size, and $\tilde{Y}_{j}$ and $\rho_{j}$ are the $j^{th}$ supervision and expected liveness score in the mini-batch. We calculate the distance between $\tilde{Y}_{j}$ and $\rho_{j}$ for our main objective function $L^{\rho}_{mse}$.

To further improve the performance, we exploit not only a regression network but also an adversarial learning technique GRL~\cite{ganin2015unsupervised}. 
Finally, our overall loss function can be formulated as follows:

\vspace{-1.0em}
\begin{equation}
\begin{aligned}
    L_{final} &= \alpha*L^{\rho}_{mse} + (1-\alpha)*L_{adv},
\end{aligned}
\label{eq:eq7}
\end{equation}
where $L^{\rho}_{mse}$ is a liveness score-based regression training loss and $L_{adv}$ is an adversarial training loss for jointly learning our livensss score-based regression neural network. $\alpha$ is a non-negative parameter to balance the importance of two losses, and we empirically set $\alpha$ to $0.5$.

\begin{table}[b]
\centering
\vspace{-1.5em}
\caption{Evaluation results for ACER (\%) in comparison with the previous methods and the proposed \textbf{PDLE} approach within the intra-dataset (OULU-NPU protocols).}
\label{tab:tab1}
\resizebox{\columnwidth}{!}{
\begin{tabular}{|c||c|c|c|c|}
\hline
\multirow{2}{*}{Method} & Protocol 1 & Protocol 2 & Protocol 3 & Protocol 4 \\ \cline{2-5} 
 & ACER(\%) & ACER(\%) & ACER(\%) & ACER(\%) \\ \hline \hline
Auxiliary~\cite{Liu_2018_CVPR} & 1.6 & 2.7 & 2.9±1.5 & 9.5±6.0 \\ \hline
CDCN~\cite{yu2020searching} & 1.0 & 1.45 & 2.3±1.4 & 6.9±2.9 \\ \hline
FaceDs~\cite{eccv18jourabloo} & 1.5 & 4.3 & 3.6±1.6 & 5.6±5.7 \\ \hline
DC-CDN~\cite{yu2021dual} & 0.4 & 1.3 & 1.9±1.1 & 4.3±3.1 \\ \hline
LMFD-PAD~\cite{LMFDPAD} & 1.5 & 2.0 & 3.4±3.1 & 3.3±3.1 \\ \hline
NAS-FAS~\cite{yu2020nasfas} & 0.2 & \textbf{1.2} & 1.7±0.6 & 2.9±2.8 \\ \hline
PatchNet~\cite{Wang_2022_CVPR} & \textbf{0} & \textbf{1.2} & 1.18±1.26 & 2.9±3.0 \\ \hline
\rowcolor{Gray} Ours & \textbf{0} & \textbf{1.2} & \textbf{0.96±1.03} &\textbf{ 0.63±1.04} \\ \hline
\end{tabular}
}
\end{table}

\vspace{-1.0em}
\section{Experiments}
\vspace{-0.7em}
\label{sec:expr}
We demonstrate the effectiveness of the proposed approach on an intra- and cross-dataset. Based on the experimental results, the characteristics of our algorithm will be discussed in this section.

\begin{table*}[]
\caption{Comparison results of cross-domain testing on MSU-MFSD (M), CASIA-MFSD (C), Replay-Attack (I), and OULU-NPU (O). PE and LE mean patch-exchange and label-encoding, respectively. \textbf{Bold} and \textit{italic} denote the best results among Res-18 and Res-50 based methods in each.}
\label{tab:tab3}
\resizebox{\textwidth}{!}{
\begin{tabular}{|c||c|cc||cc||cc||cc||}
\hline
\multirow{2}{*}{Method} & \multirow{2}{*}{Network} &\multicolumn{2}{c||}{O\&C\&I to M} & \multicolumn{2}{c||}{O\&M\&I to C} & \multicolumn{2}{c||}{O\&C\&M to I} & \multicolumn{2}{c||}{I\&C\&M to O} \\ \cline{3-10} 
 & & \multicolumn{1}{c|}{HTER(\%)} & AUC(\%)& \multicolumn{1}{c|}{HTER(\%)} & AUC(\%) & \multicolumn{1}{c|}{HTER(\%)} & AUC(\%) & \multicolumn{1}{c|}{HTER(\%)} & AUC(\%) \\ \hline \hline
NAS-FAS~\cite{yu2020nasfas} & NAS & \multicolumn{1}{c|}{19.53} & 88.63 & \multicolumn{1}{c|}{16.54} & 90.18 & \multicolumn{1}{c|}{14.51} & 93.84 & \multicolumn{1}{c|}{13.80} & 93.43 \\ \hline
NAS-FAS w/ D-Meta ~\cite{yu2020nasfas} & NAS & \multicolumn{1}{c|}{16.85} & 90.42 & \multicolumn{1}{c|}{15.21} & 92.64 & \multicolumn{1}{c|}{11.63} & 96.98 & \multicolumn{1}{c|}{13.16} & 94.18 \\ \hline
DRDG~\cite{georgecvpr2021} & DenseNet & \multicolumn{1}{c|}{12.43} & 95.81 & \multicolumn{1}{c|}{19.05} & 88.79 & \multicolumn{1}{c|}{15.56} & 91.79 & \multicolumn{1}{c|}{15.63} & 91.75 \\ \hline
ANRL~\cite{ANRL} & - & \multicolumn{1}{c|}{10.83} & 96.75 & \multicolumn{1}{c|}{17.83} & 89.26 & \multicolumn{1}{c|}{16.03} & 91.04 &
\multicolumn{1}{c|}{15.67} & 91.90 \\ \hline
LMFD-PAD~\cite{LMFDPAD} & Res-50 &\multicolumn{1}{c|}{10.48} & 94.55 & \multicolumn{1}{c|}{12.50} & 94.17 & \multicolumn{1}{c|}{18.49} & 84.72 & \multicolumn{1}{c|}{12.41} & \textit{94.95} \\ \hline
DBEL~\cite{DBEL} & Res-50 &\multicolumn{1}{c|}{8.57} & 95.01 & \multicolumn{1}{c|}{20.26} & 85.80 & \multicolumn{1}{c|}{\textit{13.52}} & \textit{93.22} & \multicolumn{1}{c|}{20.22} & 88.48 \\ \hline
HFN+MP~\cite{cai2022learning} & Res-50 & \multicolumn{1}{c|}{\textit{5.24}} & 97.28 & \multicolumn{1}{c|}{\textit{9.11}} & 96.09 & \multicolumn{1}{c|}{15.35} & 90.67 & \multicolumn{1}{c|}{\textit{12.04}} & 94.26 \\ \hline \hline
CAFD~\cite{CAFD} & Res-18 &\multicolumn{1}{c|}{11.64} & 95.27 & \multicolumn{1}{c|}{17.51} & 89.98 & \multicolumn{1}{c|}{15.08} & 91.92 & \multicolumn{1}{c|}{14.27} & 93.04 \\ \hline
SSDG-R~\cite{Jia_2020_CVPR_SSDG} & Res-18 & \multicolumn{1}{c|}{7.38} & 97.17 & \multicolumn{1}{c|}{10.44} & 95.94 & \multicolumn{1}{c|}{11.71} & 96.59 & \multicolumn{1}{c|}{15.61} & 91.54 \\ \hline
SSAN-R~\cite{ssan} & Res-18 & \multicolumn{1}{c|}{6.67} & 98.75 & \multicolumn{1}{c|}{\textbf{10.00}} & \textbf{96.67} & \multicolumn{1}{c|}{8.88} & 96.79 & \multicolumn{1}{c|}{13.72} & 93.63 \\ \hline
PatchNet~\cite{Wang_2022_CVPR} & Res-18 & \multicolumn{1}{c|}{7.10} & 98.46 & \multicolumn{1}{c|}{11.33} & 94.58 & \multicolumn{1}{c|}{13.40} & 95.67 & \multicolumn{1}{c|}{11.82} & 95.07 \\ \hline \hline
Ours w/o PE\&LE & Res-18 & \multicolumn{1}{c|}{10.83} & 94.58 & \multicolumn{1}{c|}{15.08} & 91.14 & \multicolumn{1}{c|}{14.50} & 93.55 & \multicolumn{1}{c|}{13.88} & 93.16 \\ \hline
Ours w/o PE & Res-18 & \multicolumn{1}{c|}{10.41} & 94.93 & \multicolumn{1}{c|}{13.59} & 91.04 & \multicolumn{1}{c|}{11.17} & 93.92 & \multicolumn{1}{c|}{12.50} & 94.35 \\ \hline
Ours w/o LE & Res-18 & \multicolumn{1}{c|}{9.58} & 94.47 & \multicolumn{1}{c|}{12.47} & 92.28 & \multicolumn{1}{c|}{12.25} & 94.55 & \multicolumn{1}{c|}{13.29} & 93.62 \\ \hline
\rowcolor{Gray} Ours & Res-18 & \multicolumn{1}{c|}{\textbf{5.41}} & \textbf{98.85} & \multicolumn{1}{c|}{10.05} & 94.27 & \multicolumn{1}{c|}{\textbf{8.62}} & \textbf{97.60} & \multicolumn{1}{c|}{\textbf{11.42}} & \textbf{95.52} \\ \hline
\end{tabular}
}
\vspace{-1.0em}
\end{table*}

\vspace{-0.5em}
\subsection{Datasets and Metrics}
\vspace{-0.5em}
\textbf{Datasets.} We employed four public datasets, OULU-NPU (labeled O)~\cite{oulunpu}, CASIA-FASD (labeled C)~\cite{casiamfsd}, Replay-Attack (labeled I)~\cite{replayattak}, and MSU-MFSD (labeled M)~\cite{msumfsd} for our experiments. OULU-NPU is a high-resolution database with four protocols for validating the improved performance on the intra-dataset. The videos of each dataset are recorded under different scenarios with various cameras and subjects, and they are used for cross-dataset testing to validate the generalization ability for testing data with unconstrained distribution shifts. 

\textbf{Evaluation Metrics.} 
We utilized average classification error rate (ACER) for the intra-dataset testing on OULU-NPU. The half total error rate (HTER) and area under curve (AUC) are measured for the cross-dataset testing protocols.

\vspace{-0.5em}
\subsection{Implementation Details}
\vspace{-0.5em}
\textbf{Primitive Data Preparation and Augmentation.}
Because the four FAS datasets are in video format, we extracted images at certain intervals. After obtaining the images, we used RetinaFace~\cite{deng2019retinaface} to detect faces, and then cropped and resized the color image to a resolution of 256$\times$256. Data augmentation, including horizontal flipping and random cropping, was used for training, and center cropping was employed for testing. 
And we empirically set $K$ to 10 for our approach after testing variant $K$ as depicted in Fig.~\ref{fig:graph1}.

\vspace{-0.5em}
\textbf{Experimental Setting.}
To train the FAS task, we used ResNet18~\cite{resnet18} as the encoder with the Adam optimizer under an initial learning rate and weight decay of 1e-4 and 2e-4, respectively, for all testing protocols. We trained the models with a batch size of 32 and a max epoch of 200, whereas decreasing the learning rate through an exponential LR with a gamma of 0.99. For the domain labels on the intra-dataset, we used the number of sessions in each protocol.

\vspace{-0.5em}
\subsection{Intra-Dataset Testing on OULU-NPU}
\vspace{-0.5em}
OULU-NPU has four protocols for evaluating the generalization ability under mobile scenarios with previously unseen sensors and spoof types. As shown in Table.~\ref{tab:tab1}, our PDLE approach presents the best performance for all protocols, and the expected liveness scores clearly validate the ability to generalize better latent embedding features. In particular, our proposed PDLE achieves the significant performance improvement for protocol 4 (unseen lighting, spoof type, and sensor type). 
The results demonstrate that the effectiveness to train a liveness score-based regression neural network using the amount of swapping as pseudo-discrete labels. Note that our proposed PDLE improves the overall ACER performance over the previous SOTA (PatchNet~\cite{Wang_2022_CVPR}) approach.

\vspace{-0.7em}
\subsection{Cross-Dataset Testing}
\vspace{-0.5em}
To evaluate our proposed method, we select three out of four datasets to train and use the remaining one for testing, denoted by $\{\cdot\&\cdot\&\cdot\}$ to $\{\bullet\}$. We compare our proposed method with the latest methods as shown in Table.~\ref{tab:tab3}. Among ResNet-18 based methods, we found that our method shows outstanding performance on testing the O\&C\&I to M, O\&C\&M to I, and I\&C\&M to O protocols, and the other protocol O\&M\&I to C displays the very competitive performance. When comparing to the ResNet-50 based method HFN+MP~\cite{cai2022learning}, our approach shows competitive performance on testing datasets O\&C\&I to M and O\&M\&I to C which contain a variety of image resolutions, and superior performance on testing O\&C\&M to I and I\&C\&M to O whose images are collected from various capture devices unlike other datasets.
By split testing on each capture device in the dataset C, we found that our method show relatively the low performance on low quality images (93.73\% AUC) compared to normal (94.79\% AUC) and high quality (96.47\% AUC) images. 
This result proves that the proposed method achieves satisfactory performance on all protocols because our liveness score-based regression network estimates probabilities of the real cues under various presentation attacks.

\vspace{-0.7em}
\subsection{Ablation Study}
\begin{figure}[htp]
  \centering
  {\includegraphics[width=\columnwidth]{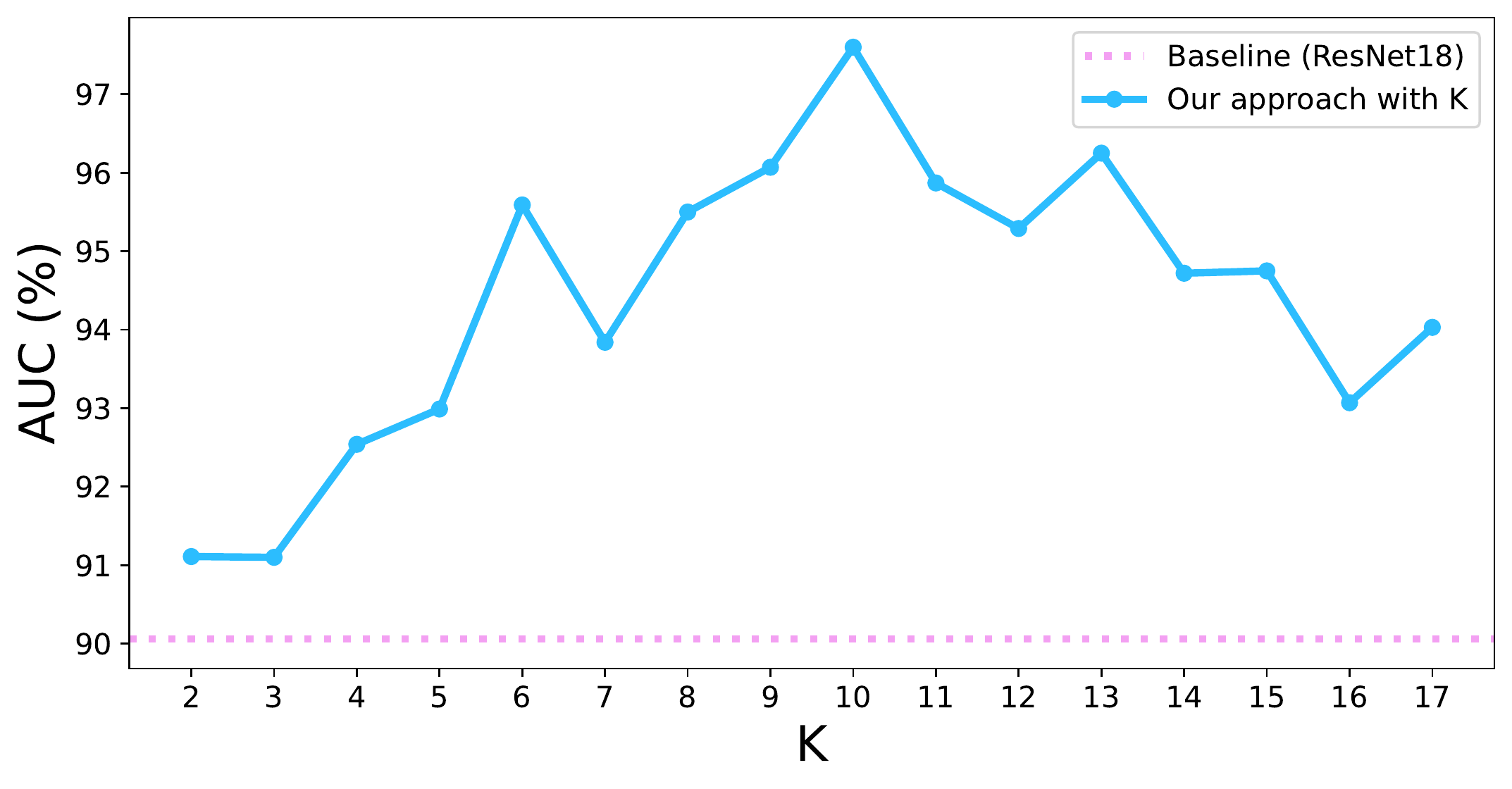}}\vspace{-1.3em}%
\caption{Ablation study on the discretized level $K$.}
\label{fig:graph1}
\end{figure}

\vspace{-0.5em}
We conducted ablation studies on cross-dataset testing to explore the contribution of each component in our method, as depicted in Table ~\ref{tab:tab3}. To analyze the effect of discretization, we separated the proposed PDLE into patch exchange (PE) and label encoding (LE). And we confirmed that each of them is the essential element for improving performance, and also observed the best performance when both were used.
In addition, we verified the influence of the pre-defined $K$ in PDLE for determining the representation power of the liveness against an input image. As shown in Fig.~\ref{fig:graph1}, we tested various values of $K$ on the O\&C\&M to I protocol to investigate the impact of $K$ on AUC. With $K$ between $2$ and $17$, our method outperforms the baseline.

\vspace{-0.8em}

\vspace{-1.7em}
\section{Conclusion}
\label{sec:conclusion}
\vspace{-0.8em}
In this paper, we have proposed the PDLE approach for training a face anti-spoofing regression model. The regression model allows the probability to estimate our liveness score. Our approach not only has the effect of a data augmentation because different labels and domains are densely exchanged, new data combinations are also created, which results in the improved domain generalization. Through our experiments, we confirm the effectiveness, robustness, and generalization of the proposed PDLE and expected liveness score.

\vspace{-0.6em}
\section{Acknowledgements}
\label{sec:acknowledgements}
\vspace{-0.8em}
This work was supported by KakaoBank Corp., and IITP grant funded by the Korea government (MSIT) (No. 2022-0-00320).
\vfill\pagebreak

\bibliography{main}
\bibliographystyle{icml2021}
\clearpage

\end{document}